\documentclass[fullpaper]{nldl}

\renewcommand{\DOI}{12345}

\usepackage[utf8]{inputenc}
\usepackage{url}
\usepackage{graphicx}
\usepackage{authblk}

\usepackage{algorithm}
\usepackage{algorithmic}
\usepackage[acronym]{glossaries}
\usepackage{caption}

\usepackage{csvsimple}
\usepackage{booktabs}
\usepackage{color,soul}

\usepackage{hyperref}

\title{Småprat: DialoGPT for Natural Language Generation of Swedish Dialogue
by Transfer Learning}
\author[1]{
Tosin Adewumi\thanks{Corresponding Author: tosin.adewumi@ltu.se\\
Presented at Northern Lights Deep Learning Conference (NLDL) 2022, Tromso, Norway.}}\author[1,2]{Rickard Brännvall}\author[1]{Nosheen Abid}\author[1]{Maryam Pahlavan}\author[1]{Sana Sabah Sabry}\author[1]{Foteini Liwicki}\author[1]{Marcus Liwicki}\affil[1]{ML Group, EISLAB, Luleå University of Technology, Sweden}\affil[2]{RISE Research Institutes of Sweden}\affil[ ]{firstname.lastname@ltu.se}

\date{\vspace{-5ex}}

\makeglossaries

\newacronym{nlp}{NLP}{Natural Language Processing}
\newacronym{ner}{NER}{Named Entity Recognition}
\newacronym{sa}{SA}{Sentiment Analysis}
\newacronym{bow}{BoW}{bag-of-words}
\newacronym{cbow}{CBoW}{continuous Bag-of-Words}
\newacronym{sltc}{SLTC}{Swedish Language Technology Conference}
\newacronym{ann}{ANN}{artificial neural network}
\newacronym{nn}{NN}{neural network}
\newacronym{lstm}{LSTM}{Long Short Term Memory Network}
\newacronym{sota}{SoTA}{state-of-the-art}
\newacronym{nlg}{NLG}{Natural Language Generation}
\newacronym{mwe}{MWE}{Multi-Word Expression}
\newacronym{sw}{SW}{Simple Wiki}
\newacronym{mt}{MT}{Machine Translation}
\newacronym{gdc}{GDC}{Gothenburg Dialogue Corpus}
\newacronym{oov}{OOV}{out-of-vocabulary}
\newacronym{gmb}{GMB}{Groningen Meaning Bank}
\newacronym{sic}{SiC}{Stockholm Internet Corpus}
\newacronym{nlu}{NLU}{natural language understanding}

\begin{document}
\nldlmaketitle

\begin{abstract}  
Building open-domain conversational systems (or chatbots) that produce convincing responses is a recognized challenge.
Recent \acrfull{sota} transformer-based models for the generation of natural language dialogue have demonstrated impressive performance in simulating human-like, single-turn conversations in English.  
This work investigates, by an empirical study, the potential for transfer learning of such models to Swedish language.
DialoGPT, an English language pre-trained model, is adapted by training on three different Swedish language conversational datasets obtained from publicly available sources: Reddit, Familjeliv and the GDC.
Perplexity score (an automated intrinsic metric) and surveys by human evaluation were used to assess the performances of the fine-tuned models.
We also compare the DialoGPT experiments with an attention-mechanism-based seq2seq baseline model,
trained on the GDC dataset.
The results indicate that the capacity for transfer learning can be exploited with considerable success.
Human evaluators asked to 
score the simulated dialogues judged over 57\% of the chatbot responses to be human-like for the model trained on the largest (Swedish) dataset.
The work agrees with the hypothesis that deep monolingual models learn abstractions which generalize across languages.
We contribute the codes, datasets and model checkpoints and host the demos on the  HuggingFace  platform.

\end{abstract}

\section{Introduction}
The introduction of Eliza, the chatbot, in the 1960s marked an epoch in the area of conversational systems \cite{weizenbaum1969computer}.
Since then, open-domain conversational systems have evolved \cite{adewumi2019conversational,adewumi2020vector}.
Advances in deep neural networks, such as the tranformer-based architectures, have brought improvements to the field \cite{devlin2018bert,radford2019language,he2020deberta}.
These models have demonstrated \acrshort{sota} performances in \acrfull{nlu} and \acrfull{nlg} tasks \cite{wang2019superglue,gehrmann2021gem}.

The advancements notwithstanding, challenges still exist with building conversational systems \cite{jurafsky2020speech, zhang2020dialogpt}.
These challenges include technical and ethical challenges\cite{javed2021understanding}.
This is more so that many of the models are originally pre-trained on English data \cite{zhang2020dialogpt}, though researchers have recently been producing multilingual versions of some of the models \cite{devlin2018mbert,NEURIPS2019_c04c19c2,xue-etal-2021-mt5}.
Some of these multilingual models, however, have been shown to have poor performance compared to models trained completely on the target language data \cite{virtanen2019multilingual,ronnqvist2019multilingual}.

In this work, we perform an empirical study of the performance of one of the recent \acrshort{sota} models, DialoGPT (medium), on various Swedish datasets of different sizes.
Their perplexity results are compared to the one we trained on the English MultiWOZ benchmark dataset.
We further compare these experiments with an LSTM-based seq2seq baseline model with the attention mechanism and
trained on the GDC dataset.
DailoGPT is an English pre-trained model for open-domain chatbots \cite{zhang2020dialogpt}.
We are not familiar with any previous published work that investigates an English pre-trained dialogue model fine-tuned to produce a different target language model.
However, there are multilingual models, which are pre-trained on unstructured text of several languages \cite{devlin2018mbert,xue-etal-2021-mt5}.
We thereby investigate how the English pre-trained model performs in \acrshort{nlg} (of dialogues) by fine-tuning on a foreign target language.

We contribute the codes\footnote{
github.com/tosingithub/gemdesk
}, datasets\footnote{We may provide the datasets or the APIs for extracting them, where applicable.}
and model checkpoints for public use and host the demos\footnote{
huggingface.co/tosin/dialogpt\_mwoz \\ huggingface.co/tosin/dialogpt\_sv
} on the HuggingFace platform.
The Swedish models are fine-tuned on extracted/crawled datasets.
The Swedish language is the official language of Sweden and is spoken by more than 8.5 million people \cite{reuter1992swedish}.
We show that generation of dialogues is possible, with reasonable performance, for a foreign, target language though the pre-training was in English.

\section{Related Work}
There are a number of pretrained models for open-domain conversational systems.
Some of them include Texar \cite{hu2018texar}, DLGnet \cite{olabiyi2019multiturn},  Meena \cite{adiwardana2020towards} and BlenderBot \cite{roller2020recipes}.
These are pretrained on dialogue datasets.
There exist, also, models pretrained on large text and adapted for conversational systems.
Examples of such models include T5 \cite{JMLR:v21:20-074} and BART \cite{lewis-etal-2020-bart}.
Another pretrained model on conversational data, DialoGPT (dialogue generative pre-trained transformer), was trained on Reddit conversations of 147M exchanges \cite{zhang2020dialogpt}.
In single-turn conversations, it achieved performance close to human in open-domain dialogues.
DialoGPT is based on GPT-2 \cite{radford2019language}.
It is an autoregressive model, which achieved \acrshort{sota} results in different \acrshort{nlp} tasks \cite{radford2019language}.

In a recent work on cross-lingual transferability \cite{artetxe-etal-2020-cross}, Artetxe et al (2020) suggest that deep monolingual models learn abstractions that generalize across languages.
This is in contrast to past hypothesis that attributes the generalization ability of multilingual models to the shared subword vocabulary used across the languages and joint training, as demonstrated for mBERT \cite{pires-etal-2019-multilingual}.
The performance of such multilingual models on low-resource languages and unseen languages are known to be poor \cite{pfeiffer2020AdapterHub, wang2021towards}.

In evaluating the performance of open-domain chatbots, it has been shown that automatic metrics, like the BLEU score, can be very poor but they are still used in some cases \cite{lundell2020conversational}.
Conversation turns per session is another metric of interest
\cite{zhou2020design}.
Perplexity is widely used for intrinsic evaluation of language models in pilot experiments and its theoretical minimum, which is its best value, is 1 \cite{adiwardana2020towards}.
Probably the best evaluation is done by human evaluators (or annotators) but this can be subjective.
The judgment of human evaluators is seen as very important, especially since humans are usually the end-users of such systems \cite{zhang2020dialogpt}.



\section{Methodology}
We used the DialoGPT medium model with 345M parameters and 24 transformer layers in this work. 
We chose this because it was reported to have the best performance (compared to its small and big versions) across a set of related tasks \cite{zhang2020dialogpt}.
The experiments were carried out on several Tesla V100 GPUs on an Nvidia DGX-1 server running Ubuntu 18.
The datasets were split in the ratio 80:10:10 for training, dev and test sets.
Multiple runs (5) per experiment were conducted and the average perplexity reported in section \ref{sec:results}.
Although one automatic metric (perplexity) was used to evaluate the models, it has been shown to correlate with another proposed human evaluation metric called Sensibleness and Specificity Average (SSA) \cite{adiwardana2020towards}.
The conversation context was set as 7 during training.
Larger contexts bring memory challenges, hence 7 appears to be a good balance for training \cite{adiwardana2020towards}.
The finetuning process involved adjusting all the parameters of the pretrained model.

Furthermore, we compare the DialoGPT experiments with a reasonable baseline model: a seq2seq model trained on the Swedish GDC dataset.
The seq2seq model is based on the LSTM architecture \cite{hochreiter1997long} and uses the attention mechanism \cite{bahdanau2015neural}.
The model has 6M trainable parameters.
A batch size of 64 is used.

\subsection{Perplexity}
Perplexity models the average predictability (i.e. minimizing the uncertainty of predicting the next token).
The lower the perplexity, the better the model performs \cite{adiwardana2020towards}.
This is used often to evaluate the language models built with n-grams of text dataset~\cite{sennrich2012perplexity}.
Perplexity, $PP$, calculates the probability $\rho$ of the test corpus, normalized by the total number of words, $N$, in the test corpus $W_{test}$.
The normalization is done by taking the Nth root of inverse of calculated probability (see Equation~\ref{eq:eq1}). 

\begin{equation}
\mathit{PP(W_{test})} = 
\sqrt[N]{ \left( \frac{1}{\rho(W_{test})} \right)}
\label{eq:eq1}
\end{equation}

\subsection{Human Evaluation}

In addition to intrinsic evaluation, the ideal model in each category of datasets was evaluated on single-turn conversations by native/near-native Swedish speakers.
The authors decided to test single-turn conversations as practiced in the original paper.
This is because the lack of long-term contextual information is still an existing problem in conversational systems \cite{zhang2020dialogpt}.
A scale of \textit{clearly human-like (4.0), somewhat human-like (3.0), not very human-like (2.0), clearly not human (1.0)} was provided.
Similarly to the original DialoGPT work \cite{zhang2020dialogpt}, we drew 30 input sentences randomly from the test set for each model and recorded their corresponding responses.

Seven human annotators then scored each conversation turn online, where the turns were
assigned randomly from one of the three models or the human ground truth.
The annotators were to use their best judgment to decide what they felt was human-like or otherwise on the four grade scale for each conversation.
For some further analysis it was decided to also aggregate annotator scores into an aggregate binary score with classes {\it not human-like} (0) and {\it human-like} (1) such that inter-annotator agreement could be defined simply as the mean of class agreement for all annotator pairs that scored the same model-question combinations.  

The English translation of part of the single-turn conversations of the familjeliv 1M+ model is available in the appendix.
They were translated using Google translate and reviewed by a Swedish native speaker.
The original Swedish conversations are also available\textsuperscript{1}.



\subsection{Byte-Pair Encoding}
A token vocabulary that includes all common words can become quite large and requires the use of an additional $<$unk$>$ special token for unknown words.
Byte-Pair Encoding (BPE) was introduced \cite{Sennrich_2016bpe} to address both these shortcomings by first identifying a base vocabulary of tokens consisting of all symbols that occur in the text and then defining merge rules based on frequencies of compound symbols to form new tokens from two tokens of the base vocabulary. The merge process proceeds until the vocabulary has attained a desired fixed size.

GPT-2 uses bytes as the base vocabulary, which forces the base vocabulary to be of size 256 while allowing all unicode characters as well as higher level subword components, basic words and common compounds to be expressed in compressed form. With some additional rules to deal with punctuation, the tokenizer of GPT-2 can handle every text without the need for the $<$unk$>$ symbol.
For GPT-2 the merge set consists of 50,000 tokens which combined with the 256 bytes base tokens and a special end-of-text token yields a total vocabulary size of 50,257.

GPT-2's tokenizer trained on English text can thus also express words in Swedish (including words with special Swedish characters å, ä and ö) as it relies on BPE, however, one can expect the compression rate to be lower as the frequency and composition of subwords are different in the two languages.
More often, it will have to rely on character level tokens.
Indeed, a comparison of the merge vocabularies of the (English) GPT-2 tokenizer and one constructed\footnote{by using the BPE tokenizer of the HuggingFace library} based on the Swedish language Gothenburg Dialogues Corpus (GDC) shows that only about 8,000 of the compound tokens are shared.  


\subsection{Datasets Used}
The authors experimented with various Swedish datasets and the English MultiWOZ.
The Swedish datasets are conversational data from Reddit (2 sizes), Familjeliv (3 sizes) and the GDC \cite{allwood2003annotations}.
The extracted data were pre-processed by removing emails, URLs, numbers and some special characters.
Table 1 summarizes the datasets.
Example conversation lines from all the Swedish datasets are available in appendix A.

\begin{table}[ht]
\caption{Summary of Datasets}
\centering
\resizebox{\columnwidth}{!}{%
\begin{tabular}{lcc}
\hline
\textbf{Dataset} &
\textbf{File Size} &
\textbf{Conversation Lines}
\\
\hline
Reddit 4K & 0.57M & 4,300
\\ 
Reddit 60K & 10.4M & 59,437
\\
\hline
Familjeliv 70K & 10.3M & 71,470
\\ 
Familjeliv 400K & 45.3M & 347,590
\\ 
Familjeliv 1M+ & 200M & 1,576,360
\\
\hline
\acrshort{gdc} & 6.6M & 108,571
\\
\hline
MultiWoZ (English) & 11M & 143,048
\\
\hline
\end{tabular}
}
\label{table:datasets}
\end{table}

\subsubsection{Reddit}
Reddit is a social discussion website with various communities or subreddits\footnote{reddit.com}.
The discussions can be very informal and contain slangs, offensive text and emojis.
The Reddit data was constructed with breadth-first search traversal, using the applicable API: PAWN.
Two sizes of the data were created and experimented with: the hot 800 topics (with about 60K conversation lines) and hot 50 topics (with over 4K conversation lines) of the Swedish subreddits.
The four Swedish subreddits from which data were extracted are \textit{sweden, svenskpolitik, swedishproblems} and \textit{stockholm}.
The sentences in the conversation were given a maximum length of 500 characters.

\subsubsection{Familjeliv.se dataset}
Familjeliv\footnote{www.familjeliv.se} is a popular website forum in Swedish. The name literally translates as family life and mainly targets adults with questions concerning pregnancy, parenthood, and domestic life hacks. It has more than a million daily visitors (2016) with lively user forum participation. The fora were scraped by automated software over the extent of a month to provide a dataset with over a million conversation turns.

\subsubsection{\acrlong{gdc}}
The \acrfull{gdc} is comprised of 360 individual dialogues transcribed from recordings of about 25 different social activities, including debates, academic seminars and situations \cite{allwood2003annotations}.
The content is somewhat different from the other Swedish corpora used in this study as it originates from real speech conversations, and contains casual language, dialect, slang and sometimes broken grammar, typical of spoken language.
It counts over 108K dialogue turns with over 1.3M tokens.

\subsubsection{MultiWOZ}
Among the many English conversational datasets available is the MultiWOZ \cite{budzianowski-etal-2018-multiwoz}.
It is a large, multi-domain and multi-task conversational dataset that has been extensively used since its creation.
It consists of more than 10,000 dialogues distributed between 70\% multi-domain
and 30\% single domain dialogues.  
It has been a standard benchmark for different dialogue problems.
There are several versions of the dataset, with each new one bringing improvements \cite{eric-EtAl:2020:LREC}.

\section{Results \& Discussion}
\label{sec:results}
Table \ref{table:res2} shows the mean perplexity results for the various datasets.
We observe a trend of decreasing perplexity score with increased dataset - as expected.
More data implies improved score.
The best perplexity score on the test set is obtained with the MultiWOZ.
This is followed by the Familjeliv size of over 1M turns while the Reddit 4K had the worst perplexity of the models trained with DialoGPT.
The English MultiWOZ model has a better perplexity compared to any of the Swedish models, though some of the latter have more conversation turns.
This observation is not surprising, as the pre-trained model was pretrained in English, though it used the Reddit-style conversation.
The seq2seq model had the worst perplexity result overall.
This should be partly because it was not pretrained. 
Its architecture is also not completely comparable to that of DialoGPT.

\begin{table}[ht]
\caption{
Mean perplexity results for the different datasets after training for 3 epochs}
\centering
\small
\begin{tabular}{lcc}
\hline
Dataset & Dev set & Test set
\\ \hline
Reddit 4K & 71.94 & 88.31 
\\ 
Reddit 60K & 65.86 & 51.70 
\\ \hline
Familjeliv 70K & 11.12 & 12.27 
\\ 
Familjeliv 400K & 7.02 & 7.44 
\\ 
Familjeliv 1M+ & 7.150 & 7.148 
\\ \hline
GDC & 29.17 & 23.95 
\\ \hline
Seq2seq-GDC & 2,864 & 2,865 
\\ \hline
MultiWOZ (English) & 6.41 & 6.21 
\\
\hline
\end{tabular}
\label{table:res2}
\end{table}

\begin{table}[ht]
\centering
\small
\caption{
Human evaluation score per model (\%)}
\begin{filecontents*}{results/score_means.csv}
annotator score,1.0,2.0,3.0,4.0
Familjeliv 1M+,0.233,0.194,0.252,0.320
Gothenburg DC,0.317,0.221,0.221,0.240
Reddit 60K,0.582,0.220,0.154,0.044
Human dialogue,0.000,0.079,0.281,0.640
\end{filecontents*}
\csvautobooktabular{results/score_means.csv}\label{table:hum_eval}
\end{table}

Table \ref{table:hum_eval} shows results from human evaluation of the presented Swedish single-turn conversations
for three of the models, with the last row including scores for the original dialogue (i.e.~ replies by real humans).
The scale is as discussed in the previous section.
The overarching target of the human evaluation was to collect a subjective opinion of how natural the simulated conversations appear to humans.
The aggregate \textit{human-like} ratio across the survey is presented in Table \ref{table:human_boots} where we see that the model trained on the large familjeliv dataset was considered human-like in 57.3\% of the assessments.
This is when the scores on the four-level scale are collapsed into binary classes for simplicity: \textit{human-like} (assigned value 1.0), which aggregates the \textit{clearly human-like} and \textit{somewhat human-like} assessments into one category, placing the other two assessments in the other class \textit{not human-like} (value 0.0).
The inter-annotator agreement scores in table \ref{table:human_boots} can then be calculated as the mean of binary class agreement across all annotator pairs that scored the same model-question combinations.

\begin{table}[ht]
\centering
\small
\caption{
Aggregate human-likeness scores (\%) and annotator agreement for each model}
\begin{filecontents*}{results/human_likeness.csv}
human-likeness,mean score,agreement
Familjeliv 1M+,0.573,0.762
Gothenburg DC,0.462,0.731
Reddit 60K,0.198,0.814
Human dialogue,0.921,0.860
\end{filecontents*}
\csvautobooktabular{results/human_likeness.csv}
\label{table:human_boots}
\end{table}

To understand better the variation in the human evaluation scores, a statistical bootstrap \cite{efron79} exercise
(i.e. sampling with replacement) was carried out and reported in table \ref{table:human_boots2}, where we note that the bootstrap mean agree well with the simple survey means for the aggregate (in table \ref{table:human_boots}) with relatively narrow standard deviations and confidence intervals (last column of table
\ref{table:human_boots2}). 
The confidence intervals for the GDC and Familjeliv1M+ overlap somewhat, while the model trained on the Reddit60K data shows considerably worse performance. This dataset had a higher occurrence of English dialogues, which may have counteracted the transfer learning into Swedish.  

\begin{table}[ht!]
\centering
\small
\caption{
Bootstrap means, standard deviations and 95\% confidence bands for the human-likeness scores}
\begin{filecontents*}{results/bootstrap_conf.csv}
human-likeness,mean,std,conf95
Familjeliv 1M+,0.571,0.049,0.47-0.66
Gothenburg DC,0.463,0.050,0.37-0.56
Reddit 60K,0.198,0.042,0.12-0.28
Human dialogue,0.921,0.028,0.86-0.97
\end{filecontents*}
\csvautobooktabular{results/bootstrap_conf.csv}
\label{table:human_boots2}
\end{table}

\begin{figure*}[ht]
\includegraphics[width=1\linewidth]{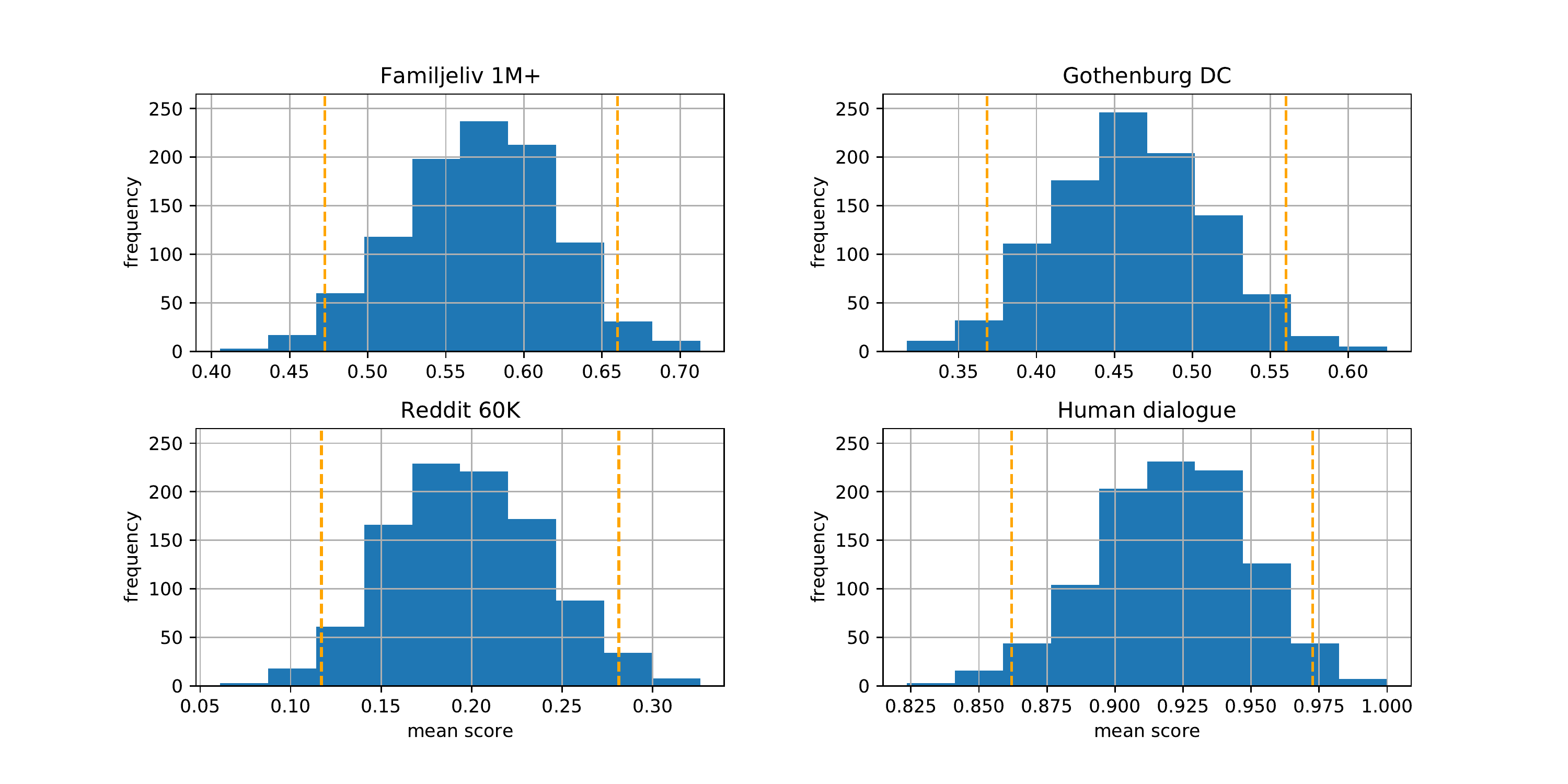}
\captionsetup{justification=raggedright,singlelinecheck=false}
\caption{Bootstrap histograms for human-likeness scores} \label{fig1}
\end{figure*}

We also note that the lower percentile for the human dialogue scores are above the higher percentile for the best performing chatbot (also visualised in Figure \ref{fig1}).
We must therefore conclude that none of the models achieved human performance.
However, this investigation seemingly agrees with the hypothesis that deep monolingual models learn abstractions that generalize across languages, as demonstrated also by \cite{artetxe-etal-2020-cross}, even though their experiments are different from those carried out in this work.
Indeed, in this work, less computational effort was needed to demonstrate this.

Demos of the English and Swedish chatbots are hosted on the HuggingFace platform\textsuperscript{3}, including the model checkpoints.
It should be pointed out that there are risks with using the models, such as producing dialogue texts that contain unwanted bias, stereotypes or offensive language.
This is a well-known, but hard to avoid problem due to the difficulty of removing such material from the underlying sources of data which the models are trained on \cite{zhang2020dialogpt}.

\section{Conclusion}
In this work, we show through an empirical study, using the recent \acrshort{sota} model DialoGPT-medium \cite{zhang2020dialogpt},
that a model pre-trained in English can be adapted by fine-tuning on a foreign language corpus such
that the performance of \acrshort{nlg} of dialogues for the 
target language can be good.
The foreign, target language in this study is Swedish.
The investigation was carried out using 3 differently-sourced Swedish datasets.
The human evaluators asked to score the simulated single-turn conversations judged that about 57\% of the chatbot responses are human-like for the model trained on the largest Swedish dataset.

Despite the advancements in deep neural networks and the demonstration of \acrshort{sota} performances in \acrshort{nlg} tasks \cite{wang2019superglue,gehrmann2021gem} by a model such as DialoGPT, challenges still exist \cite{zhang2020dialogpt}.
Hence, future work may involve attempts to understand the mathematics of languages/linguistics and their relatedness,
comparison of the performance observed here with a model pretrained on Swedish data and the possibility of using a Swedish tokenizer on the model pretrained on Swedish data.
It may also involve finding better decoding schemes than existing ones.



\bibliographystyle{abbrv}
\bibliography{references}
\printglossary[type=\acronymtype]
\section*{Appendix}
\subsection*{A - Examples from the various datasets}

\subsubsection*{Reddit examples}

\begin{quote}
"Om man tycker att munskydd är fånigt, tycker man då att ansiktsmask är ansiktslarv?"\\
"Du har en liten poäng, kylkedjan gör att det kräver mer planering och det går lite långsammare. Men det stora problemet med dessa vaccin är inte här i Sverige utan i länder som är fattiga och inte har"\\
"*Det kan jämföras med andra europeiska länder som övervakat förekomsten och som haft en tidigare spridning av denna variant där förekomsten utgör upp till  procent av fallen.* Det kan ju också jämf"\\
"Det är nu en kamp mellan vaccineringen och brittisk covid, får vi en tredje våg eller inte."\\
"Mycket lovande siffror, ska bli intressant att se SÄBO siffror om ett par veckor, det borde vara mkt låga antal då."
\end{quote}

\subsubsection{Familjeliv examples on the proper use of the word mother's day.}

\begin{quote}
"Varför? Det heter MORS DAG och FARS DAG, inte morsdag och farsdag. Det uttalas isär, inte ihopsatta som  torsdag . Så varför detta envisa ihopskrivande? Folk särskriver till höger och vänster men skriver ihop sånt som inte ska skrivas ihop. Varför?!Leo",

"Nja, man får ju skapa nya sammansatta ord på svenska så morsdag är en acceptabel konstruktion. Det är alltså inte dialekt utan ett sammansatt ord som verkar vara vanligare i vissa delar av landet.",

"Ja. Jag är också för dialekter. Men jag är inte för att skriva på dialekt. Skriftspråket är likadant var än i Sverige du befinner dig.",

"Ja, precis. Inläggen från Butterscotch och Pandamamma79 var faktiskt ganska dryga. Då kan jag lika gärna hävda att de som säger mors dag säger  fel . Det var inte meningen att vara dryg - jag menade bara att det faktiskt är bra att man får lära sig nya saker ibland.",

"Morsdag kanske kan bli ett nytt ord, men med någon annan innebörd då? Mors dag finns ju som etablerat begrepp (hur man uttalar det lägger jag mig inte i, men i skrift tycker jag att det är rimligt att alla skriver likadant).",

"Ok, tack ska du ha, jag ber om ursäkt för missförståndet och att jag blev lite putt i så fall.Men nu är ju morsdag inte ett nytt ord, det har använts mycket länge. Jag tycker att det inte spelar så stor roll om namnen inte är identiska i olika delar av landet.",

"Jag önskar alltså inte att alla i hela Sverige ska säga och skriva morsdag, men där begreppet är etablerat och till och med självklart finns det ingen anledning att ändra.Dessutom: För mig är morsdag INTE  mors dag .",

\end{quote}

\subsubsection{\acrshort{gdc} examples}

\begin{quote}

"ja men vi hade haft fest så vi var väl lite i i gasen så ja det ska jag göra sade jag"\\
"sade du det"\\
"... ja det sade jag så sen sen satte jag på den där skivan det var ett sådant där ackompanjemang till den där vi talade om den där biten när man har ... san francisco"\\
"..."\\
"ja"\\
"apropå san francisco så fick vi fick vi den hör skivan så gick vi upp och så sjöng jag till så så spelade vi upp den sen ja det lät lite knasigt men det var så löjligt men eh det är en tjusig melodi"\\
"ja"\\
"ja men det är en tjusig stad också"\\
"ja det är nog det va"\\
"ja"\\
"då har jag eh tror jag äntligen fått det hela och fungera"\\
"nejdå det var trevligt"
    
\end{quote}

\subsection*{B - Conversations with the Familjeliv 1M+ model}

\footnotesize{Prompt: We contacted the principal and she said they didn't do that. Although our principal doesn't seem very co mpetent, surely... Maybe we should contact the school counselor ourselves...\\
BOT: I've spoken to the headmaster's boss, but she doesn't have any free hours at the school. I'll try to reach h er on Monday.\\

Prompt: It turns out that the girl smeared powder into her son's hair and painted him in the face with mascara. \\BOT: Then he got angry and shouted that he was going to report this to the police.\\

Prompt: All rear-facing chairs are safer than turning one over for a young child. And then they've gone through the same tests to be sold.\\
BOT: the only thing they do anyway is drive a chair. Is it really worth it?\\

Prompt: Probably wouldn't have been able to keep your mouth shut in front of them either. It's okay to get angry with both the parents and their children when she hurts your child.\\
BOT: I love my children and do everything I can to make them feel good and have a good time.\\

Prompt: So I think merino wool can work. I think the usual wool itched a little, because the baby wasn't completely satisfied before she fell asleep...\\
BOT: I'm also afraid it's going to be too much mess on the clothes, having a hard time buying second-hand and can't really afford to castrate them.\\

Prompt: I have worked in healthcare for 10 years and have worked 7 Christmases so far. So now it's my turn to be free.\\
BOT: The first year I have only been on sick leave for 3 weeks and it was changed to 3 days a week. But I don't have a job to go back to.\\

Prompt: I think it's perfectly OK to let your child become a child model provided the child enjoys the camera and thinks it's fun.\\
BOT: We've had cameras that have been known to be good, and it's worked great.\\

Prompt: Double-check with the social insurance fund so you don't get caught for benefit fraud.\\
BOT: Thanks for the tip! I have already been in contact with fk and they would send me forms that I will submit.\\

Prompt: I'm afraid if he goes over, he's going to be that flamboyant again who's going to assert himself with new classmates, and then he's safe in the class he's in. Anyone have any wise thoughts on this?\\
BOT: He loves being in kindergarten and therefore enjoys it. However, he eats poorly and does not want to eat anything other than water and juice.\\

Prompt: In many countries, babies are given to e.g. day mother, grandmother, aunt... because parental insurance is not as generous as ours.\\
BOT: Then even if it were still good ownership, it increases the cost of a single person to make ends meet.
YOU: so we can go through it together ...
BOT: but I still want to look around a little before I get into it, to see what to expect.\\

Prompt: there are no parking facilities where as much as I know\\
BOT: You have no obligation to report anything. It is not illegal to park in a family car park.\\

Prompt: no I don't think we can make ourselves independent of nature or mo re independent of that nature\\
BOT: What exactly do you think? That there are also plants that can produce man in this way?\\

Prompt: but you won't get into the teacher's college later\\
BOT: No, but I can go in because of the training.\\

Prompt: but she can get evil too\\
BOT: And what should I do about it?\\

Prompt: uh we try flying first and we'll see\\
BOT: I've never been to the airport and don't know how it works.\\

Prompt: well then it counts the different assets as natural products then what about manure should it count you would count it as a natural resource\\
BOT: Natural products are eaten to increase the greenhouse gas in the body, which in turn causes the burning of natural tissues in nuclear power.\\
}
\end{document}